\documentclass[preprint,12pt]{elsarticle}
\usepackage{geometry} 
\usepackage{graphicx}
\usepackage{color}
\usepackage{amsmath,amsfonts,amssymb,bm,mathtools}
\usepackage{xcolor}
\usepackage{lineno}
\usepackage{subcaption}
\usepackage{booktabs}
\usepackage{siunitx}
\usepackage{algorithm,algpseudocode}
\usepackage{hyperref}
\usepackage[capitalize,nameinlink]{cleveref}
\usepackage{comment}
\usepackage{multirow}

\begin{document}

\title{Total Uncertainty Quantification in Inverse PDE Solutions Obtained with Reduced-Order Deep Learning Surrogate Models}
\author[UIUC]{Yuanzhe Wang}
\author[UIUC,PNNL]{Alexandre M. Tartakovsky\corref{mycorrespondingauthor}}
\cortext[mycorrespondingauthor]{Corresponding author}
\ead{amt1998@illinois.edu}

\address[UIUC]{Department of Civil and Environmental Engineering, University of Illinois Urbana-Champaign, Urbana, IL 61801}
\address[PNNL]{Pacific Northwest National Laboratory, Richland, WA 99352.}

\begin{abstract}

We propose an approximate Bayesian method for quantifying the total uncertainty in inverse PDE solutions obtained with machine learning surrogate models, including operator learning models. The proposed method accounts for uncertainty in the observations and PDE and surrogate models. First, we use the surrogate model to formulate a minimization problem in the reduced space for the maximum a posteriori (MAP) inverse solution. Then, we randomize the MAP objective function and obtain samples of the posterior distribution by minimizing different realizations of the objective function. We test the proposed framework by comparing it with the iterative ensemble smoother and deep ensembling methods for a non-linear diffusion equation with an unknown space-dependent diffusion coefficient. Among other problems, this equation describes groundwater flow in an unconfined aquifer. Depending on the training dataset and ensemble sizes, the proposed method provides similar or more descriptive posteriors of the parameters and states than the iterative ensemble smoother method. Deep ensembling underestimates uncertainty and provides less informative posteriors than the other two methods.
  
\end{abstract}

\maketitle

\section{Introduction} 

The model's parameters are commonly found by solving the inverse partial differential equation (PDE) problem. Given that most inverse problems are ill-posed, quantifying uncertainty in inverse solutions is important. 
Surrogate models, including machine-learning-based models, are often used to reduce the computational cost of solving inverse problems. In this work, we aim to quantify the total uncertainty in the inverse solutions including the uncertainty in the surrogate model.  

 Traditionally, uncertainty in inverse solutions is quantified using the Bayesian framework, where the \emph{posterior} distribution of states and parameters is defined in terms of the likelihood function and the prior distribution of the parameters. Markov Chain Monte Carlo (MCMC) and its variants such as Hamiltonian Monte Carlo (HMC) are the gold standard for sampling posteriors because these methods are proven to converge to the exact Bayesian posterior \cite{luengo2020survey}. However, the convergence rate of these methods is negatively affected by the problem dimensionality (the curse of dimensionality or CoD). In addition, stiff constraints in the problem (i.e., small variances of the model and measurement errors) can cause the covariance matrix of the posterior distribution to become ill-conditioned further decreasing the convergence rate \cite{langmore2023hmcillcond, neal2011mcmc, betancourt2017conceptual,zong2023randomized}. Approximate methods, including particle filters, transport maps \cite{parno2018transport}, and the ABC methods \cite{klinger2018pyabc}. 
However, most of these methods suffer from the curse of dimensionality (CoD), i.e., the number of required samples exponentially increases with the number of parameters. Machine learning (ML) surrogate models can reduce the computational time of MCMC methods, specifically, the likelihood evaluation time \cite{marzouk2009dimensionality,LI2020JCP}, but this is usually insufficient for addressing the CoD.
 
The parameter and state estimates correspodning to the maximum of the posterior distribution can be found by solving the so-called MAP (maximum a posteriori) PDE-constrained minimization problem. The disadvantage of MAP formulation is that it does not provide the uncertainty estimates and its cost increases as $N^3$, where $N$ is the number of parameters \cite{Yeung2022WRR}. The randomized MAP method  \cite{wang2018rmap} was proposed to quantify uncertainty in the MAP solution. In this approach, the objective function is randomized, and the samples are obtained using the Monte Carlo (MC) approach, i.e., by solving the minimization problem for different realizations of the objective function.  The advantage of the MC approach is that it does not depend on the number of parameters, i.e., it does not suffer from the CoD. However, the MC error decreases inverse proportionally to the square root of the number of samples. The high cost of obtaining MAP samples creates a challenge in applying the randomized MAP method to high-dimensional problems. 

Here, we use an unconstrained MAP formulation substituting the PDE constraint with the surrogate model. The approximate Bayesian method is formulated by randomizing the objective function, and the posterior samples are obtained by solving the resulting stochastic minimization problem using MC sampling. Surrogate models are subject to errors. Therefore, replacing a PDE model with the surrogate model introduces additional uncertainty in the inverse solution. Our approach accounts for the \emph{total} uncertainty including the surrogate model uncertainty. To account for the surrogate model uncertainty, the likelihood must be defined as a function of the surrogate model parameters as well as the PDE parameters.  Therefore, the likelihood needs to be marginalized before it can be evaluated. For surrogate models with many parameters, this likelihood marginalization is computationally unfeasible. This presents a challenge for MCMC and other sampling methods requiring repetitive likelihood evaluation. The advantage of our approach is that it is likelihood-free, i.e., it does not require the likelihood evaluation.  

Our approach is agnostic to the surrogate model choice but benefits from using a differentiable surrogate model (i.e., models can be analytically differentiated with respect to unknown parameters or differentiated using automatic differentiation), which is critical for efficiently solving the MAP minimization problem. The examples of differentiable surrogate models include Fourier neural operators (FNO) \cite{li2020fourier}, deep operator networks (DeepONets) \cite{lu2021learning}, graph neural operators (GNO) \cite{li2020neural},   and the reduced-order surrogate models such as the Principal Component Analysis Network (PCA-Net) \cite{bhattacharya2021model} and  Karhunen-Loève (KL)  deep neural network (KL-DNN) \cite{Wang2024CMAME} models.  The majority of the computational cost in our approach is related to solving many minimization problems, and the cost of solving each minimization problem depends on its dimensionality, i.e., the number of unknown parameters and states. Here, we employ the KL-DNN model to reduce the dimensionality and computational cost of UQ. 

This work is organized as follows. The unconstrained MAP problem in the reduced space is formulated in Section \ref{sec:MAP}. The approximate Bayesian model is presented in Section \ref{sec:Bayes}. The application of the model to the inverse non-linear diffusion equation is discussed in Section \ref{sec:Darcy}. Conclusions are given in Section \ref{sec:conclusions}.

\section{Latent-space MAP formulation of the inverse solution}\label{sec:MAP}

Consider an inverse partial differential equation problem
\begin{equation}\label{eq:PDE}
\mathcal{L}(u(\bm{x},t);y(\bm{x})) = 0, 
\end{equation}
subject to appropriate initial and boundary conditions. Here, $\mathcal{L}$ is the known differential operator, $u(\bm{x},t)$ is the state variable, and $y(\bm{x})$ is the unknown space-dependent parameter field. 

Our objective is to estimate $u(\bm{x},t)$ and $y(\bm{x})$, including uncertainty bounds, given some measurements of $u$ and, possibly, $y$, using a surrogate model mapping  $y(\bm{x})$ to $u(\bm{x},t)$. While any surrogate model can be used, we formulate the problem for reduced-order surrogate models that map a reduced space of $y(\bm{x})$ to the reduced space of $y$. Such surrogate models include the KL-DNN and PCA-Net models, which use the linear KL and PCA transformations, respectively. It is also possible to use non-linear autoencoder models, e.g., variational autoencoders. 

We assume that state and parameter fields allow reduced-order representations, $u(\bm{\bm{x}},t)  \approx \hat{u} \left (\bm{x}, t, \bm{\eta} \right )$ and $y(\bm{\bm{x}})  \approx \hat{y} \left (\bm{x},\bm{\xi} \right )$, where $\bm\eta$ and $\bm{\xi}$ are the vectors of parameters defining $u$ and $y$ in the respective latent spaces. Furthermore, we assume that the dimensionality of $\bm\eta$ and $\bm{\xi}$ is smaller than the dimensionality of $u(\bm{x},t)$ and $y(\bm{x})$ (i.e., the number of numerical elements or grids) in the numerical solution of Eq \eqref{eq:PDE}. In general, the relationship between $\bm\eta$ and $\bm{\xi}$ is non-linear and, in the PCA-NET and KL-DNN methods, is modeled with DNNs as 
$
\bm{\eta}(\boldsymbol{\xi}) \approx \mathcal{NN}(\boldsymbol{\xi},\boldsymbol{\theta}),
$
where $\boldsymbol{\theta}$ is the collection of the DNN parameters. This DNN is trained using a labeled dataset  $\bm{D}'=\{ y_i(\bm{x}) \rightarrow u_i(\bm{x},t) \}_{i=1}^{N_\text{train}}$. Here, we assume that the inverse maps $\bm \xi  = \hat{y}^{-1}(y(\bm{x}))$ and $\bm \eta  = \hat{u}^{-1}(u(\bm{x},t))$ exist. Then, the $\bm{D}'$ dataset can be reduced to the latent space dataset 
$\bm{D} = \{ \bm \xi_i \rightarrow \bm\eta_i \}_{i=1}^{N_\text{train}}$, which allows training the DNN as  
\begin{equation} \label{eq:loss_kldnn}
   \bm\theta^* = \min_{\bm\theta} \left[ L_F(\bm\theta)  = \frac{1}{\sigma^2_\eta}  \sum_{i=1}^{N_\text{train}} ||\mathcal{NN}(\boldsymbol{\xi}_i;\boldsymbol{\theta}) - \boldsymbol{\eta}_i ||^2_2 + \frac{1}{\sigma^2_\theta}  ||\boldsymbol{\theta}||^2_2 \right], 
\end{equation}
where $ L_F(\bm\theta)$ is the loss function, $||\cdot||_2$ is the $\ell_2$ norm, the last term in $ L_F(\bm\theta)$ is the regularization term, $\sigma^2_\eta$ is the variance of the DNN model error and $ \sigma^2_\theta$ is the variance of the $\bm\theta$ prior distribution. The ratio $\sigma^2_\eta / \sigma^2_\theta$ is known as the regularization coefficient. A reduced-order surrogate model can be formulated as 
\begin{equation}\label{eq:surrogate_model}
  u(\bm{x},t) \approx \hat{u} \left (\bm{x}, t; \mathcal{NN}( \boldsymbol{\xi};\boldsymbol{\theta}^*  ) \right ). 
\end{equation}
 It is important to note that $\hat{u}$ can be differentiated with respect to  $\bm{\eta}$ and $\bm\xi$ analytically or using AD, i.e., $\hat{u}$ is a differentiable model. 

With this general definition of the reduced-order surrogate model, we can formulate the inverse MAP solution in the reduced space of $\bm{\xi}$. 
We assume that $N_s^y$ measurements of $y$, 
$\boldsymbol{y}^s = [ y_1^s,..., y_{N_s^y}^s ]^T$, at locations 
$\boldsymbol{x}^y = [ x_1^y,..., x_{N_s^y}^y ]^T$
and 
$N_s^u$ measurements of $u$, 
$\boldsymbol{u}^s = [ u_1^s,..., u_{N_s^u }^s ]^T$, at locations (in space-time)
$(\bm{x}^u,t^u) = [ (x_1^u,t_1^u),..., (x_{N_s^u}^u, x_{N_s^u}^u) ]^T$
are available. Then, we formulate the inverse solution as
\begin{align}\label{eq:inv_min}
       \bm\xi^*=\min_{\bm\xi} L_I(\boldsymbol{\xi},\bm\theta^*),
\end{align}
where the loss function is defined as
\begin{align}
       L_I(\boldsymbol{\xi},\bm\theta) = 
      \gamma || \hat{\bm{u}}^s (\mathcal{NN}(\boldsymbol{\xi};\boldsymbol{\theta}) )
       - \bm{u}^s ||_2^2
     +  
     \gamma_y  || \hat{\bm{y}}^s (\boldsymbol{\xi})
       - \bm{y}^{s} ||_2^2     
       +\gamma_I \|\boldsymbol{\xi}\|_2^2.
       \label{eq:loss_inv_y}
\end{align}
Here, 
$$\hat{\bm{u}}^s (\bm\xi,\bm\theta ) = [\hat{u} (x_1^u, t_1^u; \mathcal{NN}(\bm\xi,\bm\theta )  ),...,
\hat{u} (x_{N_s^u}^u, t_{N_s^u}^u; \mathcal{NN}(\bm\xi,\bm\theta ) )]^\text{T}
$$
and
$$\hat{\bm{y}}^s (\boldsymbol{\xi}) = [\hat{y} (x_1^y,\boldsymbol{\xi} ),...,
\hat{y} (x_{N_s^y}^y,\boldsymbol{\xi} )]^\text{T}
$$ 
are the vectors of the $\hat{u}$ and $\hat{y}$ estimates of the $u$ and $y$ measurements, correspondingly, and
$\gamma_I$ is the regularization coefficient in the $\ell_2$ regularization term.  In the Bayesian interpretation, the weights $\gamma_h$ and $\gamma_y$ are inversely proportional to the variance of the errors in the $\bm{u}^{s}$ and $\bm{y}^{s}$ measurements, respectively. 

Once $\bm\xi^*$ is computed, the estimate of $\bm{y}$ and $\bm{u}$ are found as $\bm{y} \approx \hat{\bm{y}}(\bm{x}; \bm\xi^*)$ and 
 $u(\bm{x},t) \approx \hat{u} \left (\bm{x}, t; \mathcal{NN}( \boldsymbol{\xi}^*;\boldsymbol{\theta}  ) \right )$, respectively.  

\section{Approximate Bayesian parameter estimation}\label{sec:Bayes}

\subsection{KL-DNN surrogate model}

So far, the model presentation was agnostic to the type of the dimension reduction operators $\hat{y}(\bm{x})$ and  $\hat{u}(\bm{x})$. In this section, we present a Bayesian parameter estimation framework for the linear operators $\hat{y}(\bm{x})$ and  $\hat{u}(\bm{x})$ defined by the Karhunen-Loeve expansions (KLEs) as in the KL-DNN PCA-NET surrogate models. Later, we formulate a Bayesian framework for non-linear operators. 

In the KL-DNN surrogate model, the state $u$ is represented by a space-time-dependent KLE as 
\begin{equation}\label{eq:truncation_error}
 u(\bm{\bm{x}},t)  \approx \hat{u} \left (\bm{x}, t, \bm{\eta} \right ) = \bar{u}(\bm{x},t) + \sum^{N_\eta}_{i = 1} \phi_i(\bm{x},t) \sqrt{\lambda_i} \eta_i,
\end{equation}
where  $ \hat{u}$ is the KLE of $u$, $\bm{\eta} = (\eta_1, ..., \eta_{N_\eta})^T$ is the vector of unknown parameters and $\bar{u}(\bm{x},t)$, $\phi_i(\bm{x},t)$, and ${\lambda_i}$ are estimated as the properties 
(the ensemble mean and the eigenfunctions and eigenvalues of the covariance) 
of a random process that can provide an accurate statistical representation of $u(\bm{x},t)$. 
This random process is constructed by sampling $y(\bm{x})$  from its prior distribution, solving the PDE \eqref{eq:truncation_error} for each sample of $y(\bm{x})$, and computing $\bar{u}(\bm{x},t)$ and $C_u(\bm{x},\bm{x}',t,t')$ as the sample mean and covariance, respectively \cite{tartakovsky2023physics}. 

The eigenvalues ${\lambda_i}$ are organized in descending order and truncated according to the desired tolerance, rtol:
\begin{equation}
  \label{eq:truncation}
  \frac
 { \sum^{\infty}_{i = N_\eta + 1} \lambda_i}
 { \sum^{\infty}_{i = 1} \lambda_i} 
 \leq \text{rtol}. 
\end{equation}

The space-dependent parameters are represented with the standard (space-dependent) KL expansion: 
\begin{equation}
  \label{eq:tdkle}
 y(\bm{x})  \approx \hat{y} \left (\bm{x}, \bm{\xi} \right ) = \bar{y}(\bm{x}) + \sum^{N_\xi}_{i = 1} \chi_i(\bm{x}) \sqrt{\beta_i} \xi_i,
\end{equation}
where  $ \hat{y}$ is the KLE of $y$, $\bm{\xi} = (\xi_1, ..., \xi_{N_\xi})^T$ is the vector of parameters, $\bar{y}(\bm{x})$ is the prior  mean, and $\chi_i(\bm{x})$ and ${\beta_i}$ are the eigenfunctions and eigenvalues of $C_y(\bm{x},\bm{x}')$, the prior covariance of $y(\bm{x})$. The $ \hat{y}$ expansion is truncated for the desired tolerance according to the criterion similar to Eq  \eqref{eq:truncation}.   

\subsection{Randomized algorithm for parameter estimation with Gaussian model error}

There are four main sources of uncertainty in the inverse solution given by the minimization problem \eqref{eq:inv_min} and \eqref{eq:loss_inv_y}, the ill-posedness of the inverse solution, measurement errors, and the PDE and surrogate model errors. The PDE model errors are due to the assumptions in the PDE model and errors in the numerical PDE solution.  The surrogate model errors result from the KLE truncation, DNN regression error, and uncertainty due to DNN training.  

The measurement uncertainty is represented in the Bayesian framework as
\begin{equation}\label{eq:umeasnoise}
        \bm{u}^s = \bm{u}_T+\bm{\epsilon}_u
\end{equation} 
and
\begin{equation}\label{eq:ymeasnoise}
        \bm{y}^\text{s}  = \bm{y}_T + \bm{\epsilon}_y,  
\end{equation}
where $\bm{u}_T$ is the vector of the ``true'' state of the system at the measurement locations,
 $\bm{y}_T$ is the vector of the true values of the parameter field at the $y$ measurement locations, and  
 $\bm{\epsilon}_u$ and $\bm{\epsilon}_y$ are the vectors of $u$ and $y$ measurement errors, respectively. 

The $u$ model uncertainty can be represented as
\begin{equation}\label{eq:umodnoise}
   \bm{u}_T = \bm{u}(\bm\xi) + {\bm\epsilon}_M  =  \hat{\bm{u}}^s (\mathcal{NN}(\bm{\xi};\bm{\theta}^*) ) + \hat{\bm\epsilon}+  {\bm\epsilon}_M,
\end{equation}
where  $\bm{u}(\bm\xi)$ is the vector of the PDE solutions at the $u$ measurement locations given $\bm\xi$, $\hat{\bm{u}}^s(\bm\xi,\bm\theta^*)$ is the vector of the surrogate model estimates of $u$ at the measurement locations, and $\hat{\bm\epsilon}$ and 
$\bm{\epsilon}_M$ are the vectors of errors in the surrogate and PDE models, respectively. 

Here, we assume that $N_\xi$ and $N_\eta$ are chosen such that the uncertainty in the KLE models due to truncation is negligible relative to the other sources of uncertainty. Therefore, there is no uncertainty in the KLE model of $y$ ( $\bm{y}_T = \hat{\bm{y}}(\xi)$ ), and the uncertainty in the KLE model of $\bm{u}(\bm\xi)$ is only due to uncertainty in the DNN model of $\bm\eta$. It is common to assume that the measurement errors are realizations of independent identically distributed (i.i.d.) random variables. Also, it is commonly assumed that   $\hat{\bm\epsilon}$ is drawn from a Gaussian distribution. 
Here, under these assumptions, we formulate the Bayes rule for the posterior distribution of $\bm\xi$ and an algorithm for approximate sampling of this distribution. 
In Section \ref{sec:nonGaussian_est}, we propose a Bayes rule and a sampling algorithm for non-Gaussian model errors. 

In the Bayesian approach, $\bm{\xi}$ is treated as a random process with the posterior distribution $P(\bm\xi | \bm{d})$, given the measurements $\bm{d} = (\boldsymbol{u}^s,\bm{y}^s)$, defined as: 
\begin{eqnarray}\label{eq:Bayesian_posterior}
P(\bm{\xi} | \bm{d}) \propto P(\bm{d} | \bm{\xi})P( \bm{\xi}),
\end{eqnarray}
where
$P(\bm\xi)$ is the prior distribution of $\bm{\xi}$ and $P(\bm{d}| \bm\xi)$  is the likelihood function. The posterior satisfies the condition $\int P(\bm\xi | \bm{d}) \text{d} \bm\xi  = 1.$

The likelihood corresponding to the Gaussian independent errors assumption is 
\begin{eqnarray}\label{eq:likelihood}
          P(\bm{d} | \bm\xi ) = \left ( \frac{1}{\sqrt{2\pi}\sigma} \right )^{N_s^u}   \left ( \frac{1}{\sqrt{2\pi}\sigma_{y^s}} \right )^{N_s^y} 
           \exp \left (-\frac{1}{2} \delta \bm{u}^\text{T} \Sigma^{-1} \delta \bm{u}  -\frac{1}{2} \delta \bm{y}^\text{T} \Sigma_y^{-1} \delta \bm{y}  \right ),
\end{eqnarray} 
where 
$\delta \bm{u}= \hat{\bm{u}}^s (\mathcal{NN}_{\bm{\eta}^*}(\boldsymbol{\xi};\boldsymbol{\theta}) )
       - \bm{u}^s,$
$\delta \bm{y} = \hat{\bm{y}}^s (\boldsymbol{\xi})
       - \bm{y}^\text{s} $,  $\Sigma = \bm{I} \sigma^2$ ($\sigma^2 =\sigma^2_{u^s}+ \sigma^2_M+ \sigma^2_{\hat{u}}$), $\Sigma_y = \bm{I}  \sigma^2_{y^s} $ ($\bm{I}$ is the identity matrix),  and $\sigma^2_{\hat{u}}$, $\sigma^2_M$,
$\sigma^2_{u^s}$, and $\sigma^2_{y^s}$ are the variances of the KL-DNN and PDE model errors and $u$ and $y$ measurement errors, respectively.
Assuming that the prior distribution of $y$ is Gaussian, the prior distribution of $\bm\xi$, the coefficients on the KLE, is
\begin{eqnarray}\label{eq:l2_prior}
        P(\bm\xi) = (\frac{1}{\sqrt{2\pi}\sigma_\xi})^{N_\xi} 
        \Pi_{i=1}^{N_\xi}
        \exp \left[
        -\frac{\xi_i^2}{2\sigma^2_\xi} 
        \right], 
\end{eqnarray}
where $\sigma_\xi^2=1$ is the prior variance of $\bm\xi$. For the weights in the loss function \eqref{eq:loss_inv_y} set to $\gamma = \sigma^{-2}$, $\gamma_y = \sigma^{-2}_{y^s}$, and $\gamma_I = \sigma_\xi^{-2} $, the solution $\bm{\xi}^*$ of the minimization problem \eqref{eq:inv_min} provide a mode of the $P(\bm{d} | \bm\xi )$ distribution.  

MCMC methods can be used to sample low-dimensional $P(\bm\xi | \boldsymbol{d})$ posteriors. Here, we focus on high-dimensional problems (systems described by parameter fields with relatively large $N_\xi$). For such systems, we propose the approximate ``randomize-then-minimize'' Bayesian approach, where the $\bm\xi$  posterior samples are found by minimizing a stochastic objective function obtained by adding random noise in the (deterministic) loss function \eqref{eq:loss_inv_y} as
\begin{eqnarray} \label{eq:inv_sto}
    L^G_I (\boldsymbol{\xi}; \bm\theta, \boldsymbol{\alpha}^{u},
    \boldsymbol{\alpha}^{y},
    \bm\beta) 
    &=& \frac{1}{2\sigma^2}\left\| \hat{\bm{u}}^s\left( \mathcal{NN}_{\bm{\eta}}(\boldsymbol{\xi};\boldsymbol{\theta})\right)-\bm{u}^s - \boldsymbol{\alpha}^{u}\right\|_2^2 \\  \nonumber
    &+&\frac{1}{2\sigma_{y^s}^2}\left\| \hat{\bm{y}}^s \left(  \boldsymbol{\xi}\right)-\mathbf{y}^{s}-\boldsymbol{\alpha}^{y}\right\|_2^2 
     + \frac{1}{2\sigma^2_\xi} \lVert \boldsymbol{\xi}-\boldsymbol{\beta}\rVert_2^2, 
\end{eqnarray}
where $\bm{\alpha}^{u} = [\alpha_1^u,...,\alpha_{N_s^u}^u ]^\text{T}$ 
is the vector of i.i.d. normal random valuables $N(0,\sigma^2)$, 
$\bm{\alpha}^{y} = [\alpha_1^y,...,\alpha_{N_s^y}^y ]^\text{T} 
$ is the vector of i.i.d. normal random valuables $N(0,\sigma_{y^s}^2)$,
and 
$\bm{\beta} = [\beta_1,...,\beta_{N_\xi} ]^\text{T} 
$ is the vector of i.i.d. normal random valuables $N(0,\sigma_\xi^2)$.  The first term in the loss function is obtained by subtracting Eq \eqref{eq:umeasnoise} from Eq \eqref{eq:umodnoise} and defining the random noise as  $\bm{\alpha}^{u}= \bm\epsilon_u-\bm\epsilon_M-\hat{\bm\epsilon}$. Since $\bm\epsilon_u$, $\bm\epsilon_M$, and $\hat{\bm\epsilon}$ are the vectors of zero-mean independent Gaussian variables, $\bm{\alpha}^{u}$ is also the vector of independent Gaussian random variables with zero mean and variance $\sigma^2 =\sigma^2_{u^s}+ \sigma^2_M+ \sigma^2_{\hat{u}}$. The second term in the loss function enforces Eq \eqref{eq:ymeasnoise}. The last term in the loss function expresses the prior knowledge. 

The samples $\{ \bm{\xi}^{(i)} \}_{i=1}^{N_\text{ens}}$ are obtained by minimizing the loss function 
$$L^G_I (\boldsymbol{\xi}; \bm\theta^*, \bm{\alpha}^{u(i)},
    \boldsymbol{\alpha}^{y(i)},
    \bm\beta^{(i)}),$$
where $\boldsymbol{\alpha}^{u(i)}$, $\boldsymbol{\beta}^{(i)}$, and $\boldsymbol{\alpha}^{y(i)}$ are the samples of the  $\boldsymbol{\alpha}^{u}$, $\boldsymbol{\beta}$, and $\boldsymbol{\alpha}^{y}$ random variables.
The remaining questions are how to 
estimate $\sigma^2_{\hat{u}}$ and, if necessary, relax the assumption of the  $\hat{\bm\epsilon}$ Gaussian distribution. The methods for sampling the posterior distribution of the DNN parameters and quantifying uncertainty in the Kl-DNN model, including $\sigma^2_{\hat{u}}$, are described in the following section. 
  
\subsection{Uncertainty in the forward KL-DNN model}
To quantify the DNN model uncertainty, two methods were proposed in \cite{Wang2024CMAME}, the deep ensembling (DE) method quantifying uncertainty due to the DNN parameter initialization in solving the minimization problem in Eq \eqref{eq:loss_kldnn} and the randomized KL-DNN (rKL-DNN) method, which also accounts for uncertainty due to the DNN and training datasets finite size resulting in approximation errors. In both approaches, $\bm\theta^*$ is treated as a random vector $\bm\Theta^*$. 

In the DE method, the minimization problem \eqref{eq:loss_kldnn} is randomly initialized and solved $N_\text{ens}$ times, yielding the ensemble of DNN parameters $\{ \bm\theta^{(i)}_{DE} \}_{i=1}^{N_\text{ens}}$, which correspond to the locations of the modes of the $\bm\theta$ posterior distribution given the Gaussian prior and likelihood functions. It should be noted that the $ \bm\theta^{(i)}_{DE}$ samples define a distribution of $\bm\Theta^*$ that is different from the Bayesian posterior distribution of $\bm\Theta^*$.  

In the rKL-DNN method, the posterior distribution is approximately sampled by minimizing the randomized $L_F$ loss function as
\begin{align}
    L_{rF}(\bm\theta) = \frac{1}{\sigma^2_\eta} \sum_{i=1}^{N_\text{train}} ||\mathcal{NN}(\boldsymbol{\xi}_i, \boldsymbol{\theta}) - \boldsymbol{\eta}^{data}_i - \boldsymbol{\alpha_i}||^2_2  + \frac{1}{\sigma^2_\theta} ||\boldsymbol{\theta} -\boldsymbol{\beta}||^2_2, 
\end{align}
where $\boldsymbol{\alpha}_i$ $(i=1,...,N_\text{train})$ and $\boldsymbol{\beta}$ are vectors of i.i.d. Gaussian random variables with zero mean and variances $\sigma_\eta^2$ and $\sigma_\theta^2$, respectively. 

Minimizing the randomized loss function for different realizations of $\{ \boldsymbol{\alpha}_i \}_{i=1}^{N_\text{train}}$, $\boldsymbol{\beta}$, and the random initial guesses of $\boldsymbol{\theta}$ produces samples $\{ \bm{\theta}^{(i)}_\text{r} \}_{i=1}^{N_\text{ens}}$ of the posterior distribution of $\boldsymbol{\theta}$. The $\sigma_\eta^2$ and $\sigma^2_\theta$ parameters are chosen such as to maximize the log predictive probability (LPP) with respect to \emph{testing} data \cite{rasmussen2006gaussian}: 

\begin{eqnarray}\label{eq:lpp}
    \mathrm{LPP} = -\sum_{i = 1}^{N_x} \sum_{j = 1}^{N_t} \Big\{ &&
    \frac{
[
     \overline{u}_\text{rKL-DNN} (\bm{x}_i,t_j;\bm{\xi}_\text{test})  - u_\text{test} ( \bm{x}_i,t_j )    
    ]^2 
    }
    {    2  \sigma^2_\text{rKL-DNN} (\bm{x}_i,t_j; \bm{\xi}_\text{test}  )   } \\ \nonumber
    &+&  \frac{1}{2}\log [2\pi \sigma^2_\text{rKL-DNN} (\bm{x}_i,t_j; \bm{\xi}_\text{test}))]
    \Big\},
\end{eqnarray} 
where 
$  \overline{u}_\text{rKL-DNN} (\bm{x}_i,t_j;\bm{\xi}_\text{test}) $
and
$\sigma^2_\text{rKL-DNN} {(\bm{x}_i,t_j;\bm{\xi}_\text{test}))}$ are the sample mean and variance of $u(\bm{x},t)$ given the parameters $\bm{\xi}_\text{test}$ computed from the rKL-DNN method as 
\begin{equation}\label{eq:KLDNNmean}
  \bar{u}_\text{rKL-DNN}(\bm{x},t; \boldsymbol{\xi}) = \frac{1}{N_\text{ens}}\sum_{i=1}^{N_\text{ens}}   
  \hat{u} \left (\bm{x}, t, \mathcal{NN}( \boldsymbol{\xi};\bm{\theta}^{(i)}_\text{r}  ) \right )
\end{equation}
and
\begin{equation}
  \sigma^2_\text{rKL-DNN}(\bm{x},t;  \boldsymbol{\xi})  = \frac{1}{N_\text{ens}-1}\sum_{i=1}^{N_\text{ens}}   [\hat{u} \left (\bm{x}, t, \mathcal{NN}( \boldsymbol{\xi};\bm{\theta}^{(i)}_\text{r}  ) \right )
  -  \bar{u}_{rKL-DNN}(\bm{x},t; \boldsymbol{\xi})]^2.
\end{equation}

\subsection{Accounting for the surrogate model uncertainty in the inverse solution}\label{sec:nonGaussian_est}
For the uncertain (random) DNN model, the likelihood given by Eq \eqref{eq:likelihood}
becomes a function of the random DNN parameter vector $\bm\Theta$,  $P(\bm{d} | \bm\xi, \bm\Theta )$, and must be marginalized as
\begin{equation}\label{eq:marg_likelihood}
 P(\bm{d},\bm{D} | \bm\xi ) = \int P(\bm{d} | \bm\xi, \bm\Theta ) P(\bm\Theta|\bm{D}) d\bm\Theta
\end{equation}
before the posterior given by Bayes's rule
\eqref{eq:Bayesian_posterior} can be sampled. Here, $\bm{d}$ is the vector of the $u$ ``field'' measurements, and $\bm{D}$ is the synthetic dataset for training the forward surrogate model.  Theoretically, the samples $\{\bm{\theta}^{(i)}_\text{r} \}_{i=1}^{N_\text{ens}}$ of the posterior distribution $P(\bm\Theta|\bm{D})$ can be used to numerically compute the marginalized likelihood in Eq \eqref{eq:marg_likelihood}. However, given the high dimensionality of $\bm{\Theta}$, such computations of the marginal likelihood (which have to be performed multiple times in MCMC and similar sampling methods) are unfeasible.   

Here, we propose a randomized algorithm for sampling the posterior distribution of $\bm\xi$, which avoids  $P(\bm{d},\bm{D} | \bm\xi )$ evaluations.  
We assume that the errors in the randomized KL-DNN model are unbiased, i.e., the KL-DNN error model in Eq \eqref{eq:umodnoise} can be rewritten as   

\begin{equation}\label{eq:umodnoise_mean}
   \bm{u}(\bm\xi) =  \bar{\bm{u}}_\text{rKL-DNN}(\boldsymbol{\xi})  + \bm{u}'_\text{rKL-DNN}(\boldsymbol{\xi},\bm\theta) 
   =  \hat{\bm{u}} \left (\mathcal{NN}( \bm{\xi};\bm{\theta}  ) \right ), 
\end{equation}
where  $\bar{\bm{u}}_\text{rKL-DNN}$ is the vector of the mean KL-DNN predictions of $u$ at the measurement locations and 
$\bm{u}'_\text{rKL-DNN} (    \bm{\xi}; \bm{\theta}     )$ 
is the vector of zero-mean fluctuations (errors). In doing this, we replace  $\hat{\bm{u}}(\bm{\xi},\bm\theta^*)$, the KL-DNN estimate of $u$ corresponding to one of modes of the $\bm\Theta$ distribution, with  $\bar{\bm{u}}_\text{rKL-DNN}(\boldsymbol{\xi})$  and the error vector $\hat{\bm\epsilon}$ with the perturbation vector defined as 
$\bm{u}'_\text{rKL-DNN}(\boldsymbol{\xi},\bm\theta) = 
\hat{\bm{u}} \left (\mathcal{NN}( \bm{\xi};\bm{\theta}  ) \right ) - \bar{\bm{u}}_\text{rKL-DNN}(\boldsymbol{\xi}).
$  
The advantage of this error model is that it allows for a non-Gaussian correlated error distribution defined by the samples $\{ \bm{u}'_\text{rKL-DNN}(\boldsymbol{\xi},\bm\theta^{(i)}_\text{r}) \}_{i=1}^{N_\text{ens}}$. Then, we can rewrite Eq \eqref{eq:umodnoise} as
\begin{equation}\label{eq:umodnoise-nG}
 \bm{u}_T =  \hat{\bm{u}}^s (\mathcal{NN}(\bm{\xi};\bm{\theta}) ) +  {\bm\epsilon}_M,
\end{equation}
and the (Gaussian) loss $L^G_I$ as  
\begin{eqnarray} \label{eq:inv_sto}
    L^{nG}_I (\boldsymbol{\xi}; \bm\Theta, \boldsymbol{\alpha}^{u},
    \boldsymbol{\alpha}^{y},
    \bm\beta) 
    &=& \frac{1}{2\sigma^2}\left\| \hat{\bm{u}}^s\left( \mathcal{NN}_{\bm{\eta}}(\boldsymbol{\xi};\boldsymbol{\Theta})\right)-\bm{u}^{\mathrm{s}} - \tilde{\boldsymbol{\alpha}}^{u}\right\|_2^2 \\  \nonumber
    &+&\frac{1}{2\sigma_{y^s}^2}\left\| \hat{\bm{y}}^s \left(  \boldsymbol{\xi}\right)-\mathbf{y}^{\mathrm{s}}-\boldsymbol{\alpha}^{y}\right\|_2^2 
     + \frac{1}{2\sigma^2_\xi} \lVert \boldsymbol{\xi}-\boldsymbol{\beta}\rVert_2^2, 
\end{eqnarray}
where $\tilde{\bm{\alpha}}^{u} = [\alpha_1^h,...,\alpha_{N_s^h}^h ]^T$ 
is the vector of i.i.d. normal random valuables $N(0,\tilde{\sigma}^2)$ and  
$\tilde{\sigma}^2 =\sigma^2_{u^s}+ \sigma^2_M$.

The samples $\{ \bm{\xi}^{(i)}_r \}_{i=1}^{N_\text{ens}}$ are obtained by minimizing the loss function 
$L^{nG}_I$:
\begin{align} 
  \bm\xi^{(i)}_r = \min_{\bm\xi }L^{nG}_I (\boldsymbol{\xi}; \bm\theta^{(i)}_\text{r}, \bm{\alpha}^{u(i)},
    \boldsymbol{\alpha}^{y(i)},
    \bm\beta^{(i)}). 
\end{align}
Then, the mean 
$\overline{y}_{\text{post}}(\bm{x})$ 
 and variance 
 $\sigma^2_{y,\text{post}}(\bm{x})$
 of the $y(\bm{x})$ posterior distribution can be computed as
 \begin{equation}\label{eq:sample_mean_y}
\overline{y}_\text{post} (\bm{x})= \frac1{N_\text{ens}} \sum_{i=1}^{N_\text{ens}}  \hat{y} ( \bm{x},\bm\xi^{(i)}_{r} )   
\end{equation}   
and 
\begin{equation}\label{eq:sample_var_y}
\sigma^2_{y, \text{post}}(\bm{x}) = \frac1{N_\text{ens}-1} \sum_{i=1}^{N_\text{ens}} ( \hat{y} (\bm{x}, \bm\xi^{(i)}_{r} ) - \overline{y}_{\text{post}}(\bm{x})    )^2.
\end{equation}  
 We term this method for computing the approximate posterior distribution of the inverse solution the \emph{randomized} inverse KL-DNN method or rI-KL-DNN. 

It can be shown that if  $\hat{\bm{u}}\left( \mathcal{NN}(\boldsymbol{\xi};\bm{\theta})\right)$ is linear in $\bm{\xi}$ and $\bm\theta$, then the proposed randomized approach converges to the \emph{exact} posterior given by the Bayes rule \cite{zong2023randomized}. \ For the non-linear surrogate model $\hat{\bm{u}}$ (as is the case for the KL-DNN model), the possible bias in the randomized algorithms can be removed using Metropolis rejection algorithms \citep{zong2023randomized}. However, it was found that the rejection rate in randomized algorithms is very low \citep{oliver1996conditioning, wang2018rmap,zong2023randomized}. Therefore, in this work, we accept all samples generated by the rI-KL-DNN algorithm.

The rI-KL-DNN algorithm can be generalized for a generic surrogate model 
$\hat{\bm{u}} ( \bm{y};\bm{\theta})$, in which case the randomized loss function becomes  
\begin{eqnarray} \label{eq:inv_sto_gen}
    \tilde{L}^{nG}_I (\bm{y}; \bm\theta, \boldsymbol{\alpha}^{u},
    \boldsymbol{\alpha}^{y},
    \bm\beta) 
    &=& \frac{1}{2\sigma^2}\left\| \hat{\bm{u}}^s\left( \bm{y};\bm{\theta}\right)-\bm{u}^{\mathrm{s}} - \boldsymbol{\alpha}^{u}\right\|_2^2 \\  \nonumber
    &+&\frac{1}{2\sigma_{y^s}^2}\left\| \hat{\bm{y}}^s -\bm{y}^{\mathrm{s}}-\boldsymbol{\alpha}^{y}\right\|_2^2 
     + \frac{1}{2\sigma^2_y} \lVert \bm{y}-\tilde{\bm{\beta}}\rVert_2^2, 
\end{eqnarray}
where $\bm{y}$ is the vectors of $y(\bm{x})$ values accurately describing $y(\bm{x})$, $\tilde{\bm{\beta}}$ is the vector of random variables, whose distribution is given by the prior distribution of $\bm{y}$ and $\sigma^2_y$ is the prior variance of $y$. 

Finally, we describe the DE approach for the inverse KL-DNN method as an extension of the DE-KL-DNN method.  In this approach, the samples $\{ \bm{\xi}^{(i)}_{DE} \}_{i=1}^{N_\text{ens}}$ are obtained by minimizing the loss function 
$L_I$:
\begin{align} 
  \bm\xi^{(i)}_{DE} = \min_{\bm\xi }L_I (\boldsymbol{\xi}; \bm\theta^{(i)}_\text{DE} ). 
\end{align}
Then, the mean 
$\overline{y}_{\text{DE}}(x)$ 
 and variance 
 $\sigma^2_{y,\text{DE}}(x)$
 of the DE $y(x)$ estimate are obtained as
 \begin{equation}\label{eq:DE_mean_y}
\overline{y}_\text{DE} (\bm{x})= \frac1{N_\text{ens}} \sum_{i=1}^{N_\text{ens}}  \hat{y} ( \bm{x},\bm\xi^{(i)}_{DE} )   
\end{equation}   
and 
\begin{equation}\label{eq:DE_var_y}
\sigma^2_{y, \text{DE}}(\bm{x}) = \frac1{N_\text{ens}-1} \sum_{i=1}^{N_\text{ens}} ( \hat{y} (\bm{x}, \bm\xi^{(i)}_{DE} ) - \overline{y}_{\text{DE}}(\bm{x})    )^2.
\end{equation}

\section{Non-linear two-dimensional diffusion equation}\label{sec:Darcy}

\subsection{Problem formulation}

\begin{figure}[H]
    \centering
    \includegraphics[height=0.80\textwidth]{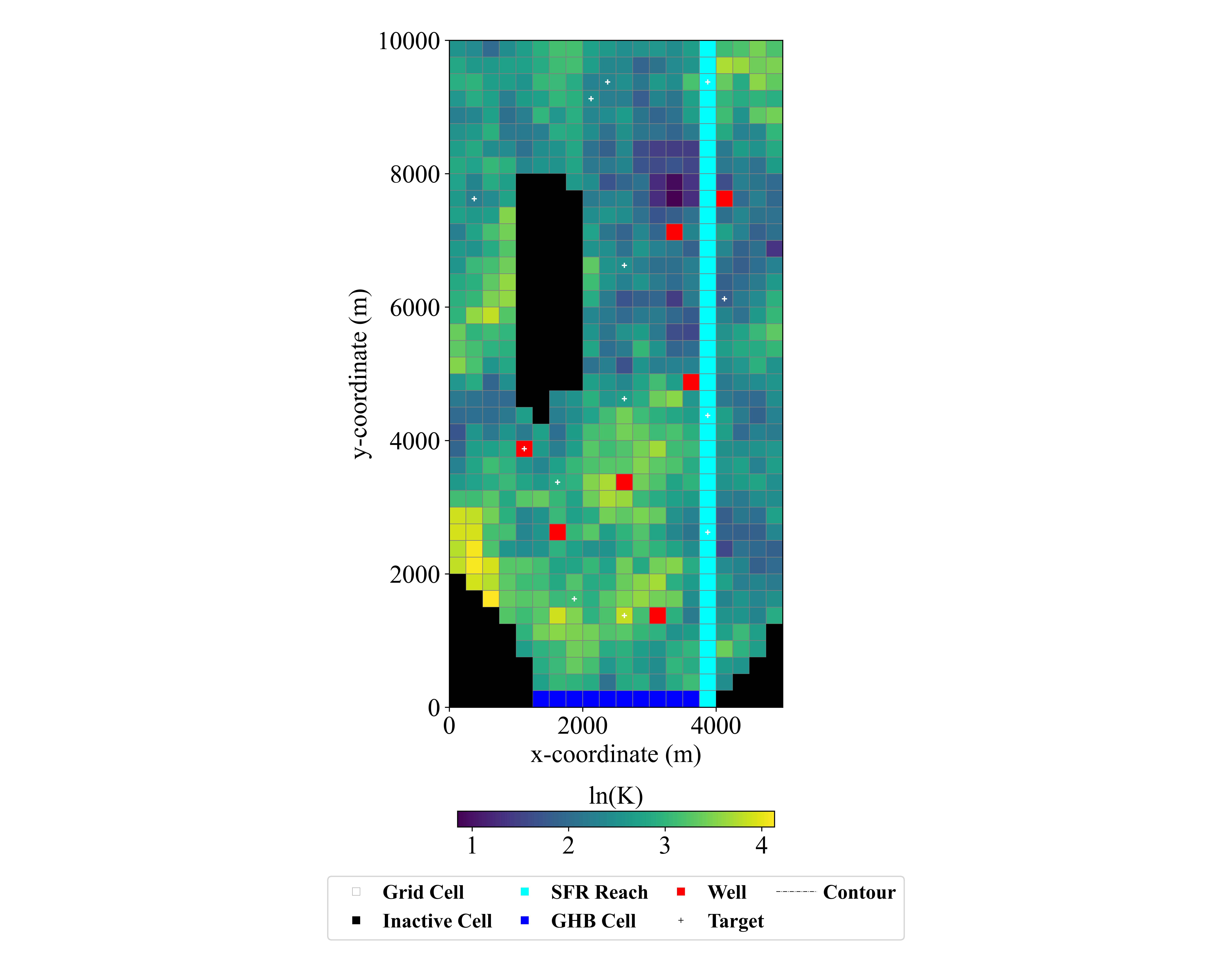}
    \caption{
    The discretized domain of the hypothetical ``Freyberg'' aquifer. The color map shows the reference log hydraulic conductivity field $y_\text{ref}(\bm{x})=\ln K_\text{ref}(\bm{x})$.
    Inactive or no-flow cells, the stream reach, general head boundary (GHB) cells, pumping wells, and observation wells are also shown.}
    \label{fig:2d_diffu_y_ref}
\end{figure}

In this section, we test the DE-KL-DNN and rI-KL-DNN methods for groundwater flow in a synthetic unconfined aquifer known as the Freyberg problem \cite{freyberg1988exercise,hunt2020revisiting}. The governing equation for this problem is non-linear and has the form:
 \begin{equation}
 S_y u(\bm{x},t) \frac{\partial u(\bm{x},t)}{\partial t}=  \nabla \cdot(K(\bm{x}) u(\bm{x},t) \nabla u(\bm{x},t) ) +  u(\bm{x},t) [f(t)+g(\bm{x},t)] 
  \label{eq:Darcy}
 \end{equation} 
where $S_y$ is the specific yield (here, assumed to be constant and known), $K(\bm{x})$ is the hydraulic conductivity, $f(t)$ is the time-dependent recharge, and $g(\bm{x},t)$ is the source term due to pumping wells and the interaction with a river. The computational domain, boundary conditions, and the discretization mesh are shown in Figure \ref{fig:2d_diffu_y_ref} and, along with the initial condition, are discussed in detail in \cite{Wang2024CMAME}. Figure \ref{fig:2d_diffu_y_ref}  also shows the reference log conductivity field $y_\text{ref} (\bm{x}) = \ln K_\text{ref} (\bm{x})$, which is generated as a realization of a correlated random field 
using the software package PyEMU \citep{white2020toward,white2016python} as described in \cite{Wang2024CMAME}. PyEMU is also used to generate $N_{train}$ realizations of $y(\bm{x})$, $\{ y^{(i)}(\bm{x}) \}_{i=1}^{N_\text{train}}$, to train the KL-DNN model.  
The reference $u$ field, $u_{ref}(\bm{x},t)$ is found by solving Eq \eqref{eq:Darcy} with $y(\bm{x})=y_{ref}(\bm{x})$ using the MODFLOW 6 (MF6) software \cite{langevin2024modflow, langevin2017documentation} as described in \cite{Wang2024CMAME}. Similarly, MF6 is used to generate $\{ u^{(i)} (\bm{x},t) \}_{i=1}^{N_\text{train}}$ fields using  $\{ y^{(i)}(\bm{x}) \}_{i=1}^{N_\text{train}}$ as inputs.

The samples  $\{ y^{(i)} = \ln K^{(i)} (\bm{x}) \}_{i=1}^{N_\text{train}}$ and $\{ u^{(i)} (\bm{x},t) \}_{i=1}^{N_\text{train}}$ are used to construct the KLEs of $y(\bm{x})=\ln K(\bm{x})$ and $u(x,t)$ and to train the DNNs $\{ \mathcal{NN}_\eta(\bm\xi;\bm\theta^{(i)}_\text{r} ) \}_{i=1}^{N_\text{ens}}$ and $\{ \mathcal{NN}_\eta(\bm\xi;\bm\theta_\text{DE}^{(i)}) \}_{i=1}^{N_\text{ens}}$ using the rKL-DNN and DE-KL-DNN methods.   
Following \cite{Wang2024CMAME}, we set $N_\xi=150$ (rtol$=0.069$), $N_\eta = 90$ (rtol$=0.00045$), $N_\text{ens} = 100$, $\sigma_{\eta}^2=10^{-8}$, and $\sigma_{\theta}^2=10^{-3}$. 

\subsection{Uncertainty in the forward surrogate model }
Here, we quantify errors and uncertainty in the KL-DNN model. 
\begin{figure}[H]
    \centering
    \includegraphics[width=0.60\textwidth]{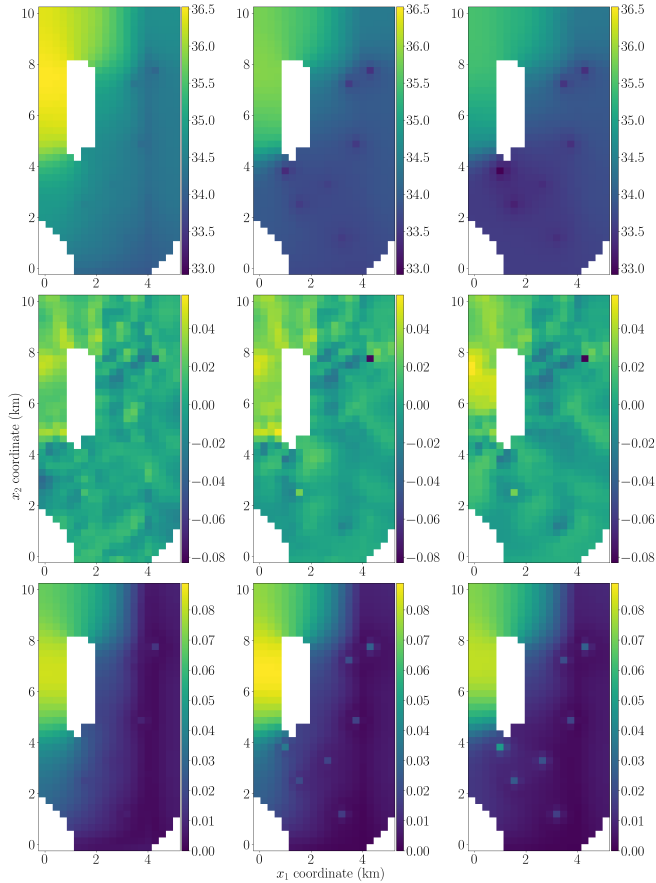}
    \caption{
The reference field $u_\text{ref}$ at times $t_1=10$, $t_2= 11$, and $t_3=12$ years  (first row).
 The point error $\bar{u}_\text{rKL-DNN}(\bm{x},t)-u_\text{ref}(\bm{x},t)$  (second row). The standard deviation $\sigma_\text{rKL-DNN}(\bm{x},t)$ 
 (third row).
 The rKL-DNN model is trained with $N_\text{train}=5000$ samples.  
The average standard deviations of $u$ at the observation locations are 0.024 at $t=t_1$, 0.023 at $t=t_2$, and 0.022 at $t=t_3$.}
    \label{fig:2d_diffu_u_comp}
\end{figure}

Figure \ref{fig:2d_diffu_u_comp} shows the reference hydraulic head field $u_\text{ref}$ at times $t_1=10$, 
$t_2= 11$, and $t_3=12$ years, the point difference $\bar{u}_\text{rKL-DNN}(\bm{x},t)-u_\text{ref}(\bm{x},t)$ (error in the mean rKL-DNN solution) and the standard deviation in the rKL-DNN solution. 
The spatial distribution of errors and standard deviations is not uniform. The maximum error and standard deviation locations coincide with the maximum $u$ location. The maximum error and standard deviation are about $0.3\%$ of the maximum $u$ value.

\subsection{Total uncertainty in the inverse solution}

In this section, we use rI-KL-DNN and DE-KL-DNN methods to obtain inverse solutions and quantify uncertainty in these solutions.
We benchmark these methods against the iterative ensemble smoother (IES) method for several $N_\text{train}$ values. 
IES seeks to minimize errors between ensemble model outputs and observations by adjusting the model parameter ensemble. The Monte Carlo specification of a prior parameter ensemble yields, upon model forward run, an ensemble of model outputs, whose residuals at observed locations and times can be used in an ensemble form of the Gauss-Levenberg-Marquardt (GLM) algorithm. This form of the GLM approximates the Jacobian of parameters and model outputs from their empirical error covariances using a linear regression model \cite{chen2013levenberg}. Successive iterations (or batch runs) of the ensemble smoother (model forward runs followed by GLM updates of the parameters) reduce the ensemble output residuals and provide samples of the estimated parameter distribution. These samples are used to compute the mean $\overline{y}_\text{IES} (\bm{x})$ and variance $\sigma^2_{y, \text{IES}}(\bm{x})$ of $y_\text{ref}(\bm{x})$.
For a fair comparison between IES and KL-DNN inverse solutions, the total number of MF6 (forward model) calls in the IES algorithm is set to $N_\text{train}$. Three IES algorithm iterations are found to be sufficient for the considered problem.

We assume that noisy measurements $\boldsymbol{u}^s$ of $u(\bm{x},t)$ are available at 25 instances of time at the 13 locations marked by crosses in Figure \ref{fig:2d_diffu_y_ref}. Reliable direct measurements of the aquifer properties are usually unavailable in real-world applications. Therefore, we do not use any measurements of $y$ to obtain the inverse solution.  We also assume that the PDE model is exact and set $\sigma^2_M=0$.
The $u$ measurements are generated as 
\begin{equation}\label{eq:umeasnoise_gen}
        \bm{u}^s = \bm{u}_{ref} + \bm{\epsilon}_u
\end{equation} 
where $\bm{u}_{ref}$ is the vector of the $u_{ref}(\bm{x},t)$ values at the measurement locations and $\bm{\epsilon}_u$ is the vector of uncorrelated zero-mean Gaussian variables with the variance $\sigma^2_{y^s}$.

\begin{figure}[H]
    \centering
    \includegraphics[width=0.80\textwidth]{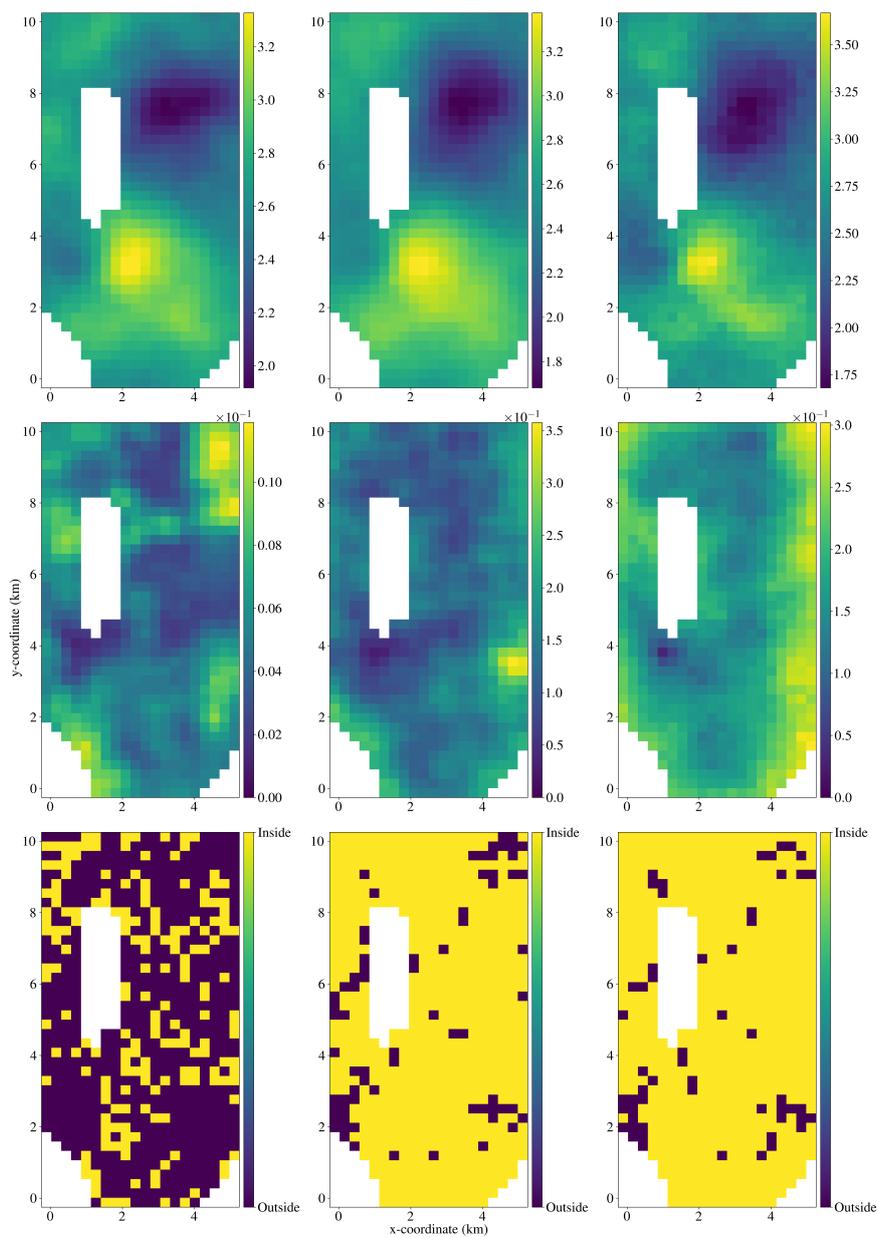}
    \caption{The estimated $y$ field with DE-KL-DNN (left column), rKL-DNN (mid column), and IES (right column). $N_\text{train} = 5000$ and $\sigma^2_{y^s} = 0.01$. The top row is the estimated $\overline{y}(\bm{x})$, the mid row is the estimated $\sigma^2_y(\bm{x})$, and the bottom row is the coverage.}
    \label{fig:2d_diffu_coverage_0_1}
\end{figure}

\begin{figure}[H]
    \centering
    \includegraphics[width=0.80\textwidth]{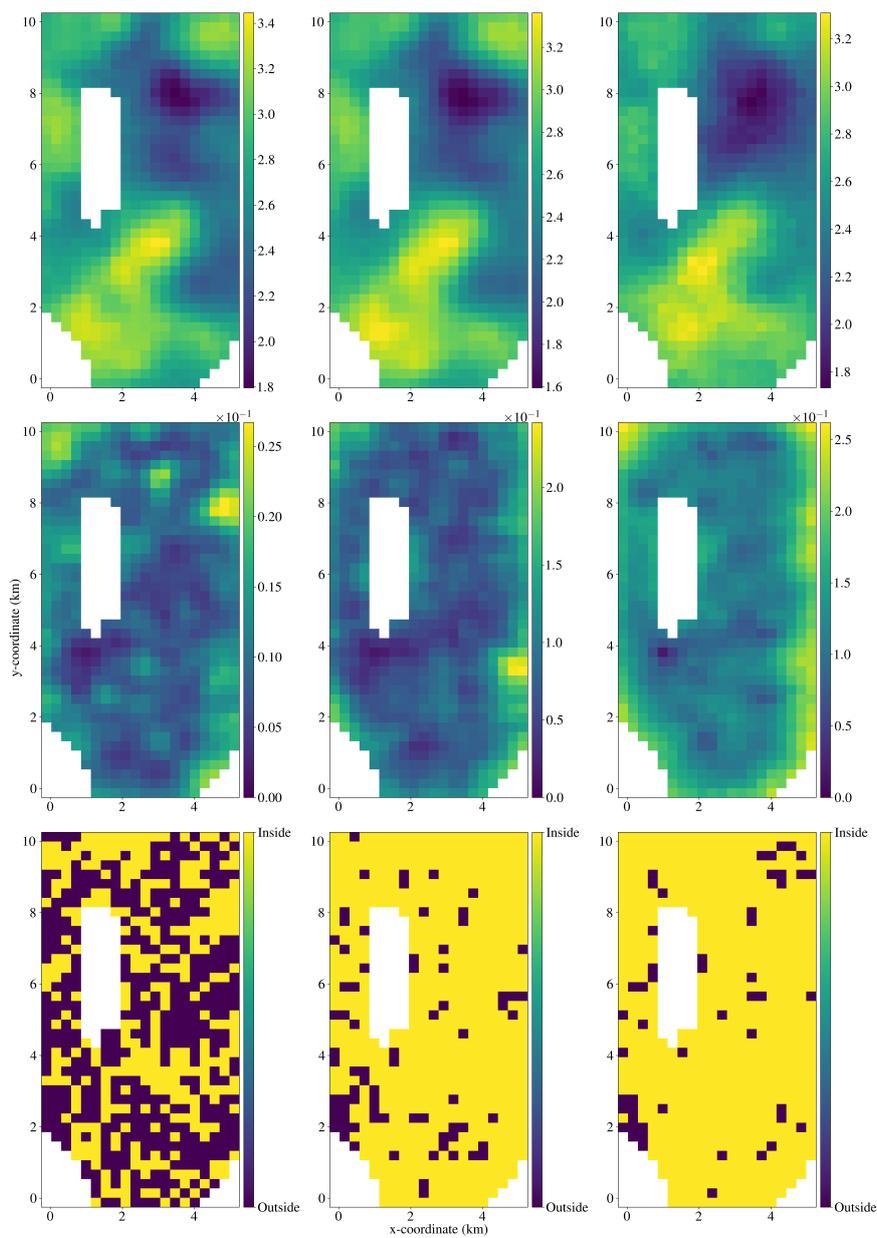}
    \caption{The estimated $y$ field with DE-KL-DNN (left column), rKL-DNN (mid column), and IES (right column). $N_\text{train} = 5000$ and $\sigma^2_{y^s} = 10^{-4}$. The top row is the estimated $\overline{y}(\bm{x})$, the mid row is the estimated $\sigma^2_y(\bm{x})$, and the bottom row is the coverage.}
    \label{fig:2d_diffu_coverage_0_01}
\end{figure}

\begin{figure}[H]
    \centering
    \includegraphics[width=0.90\textwidth]{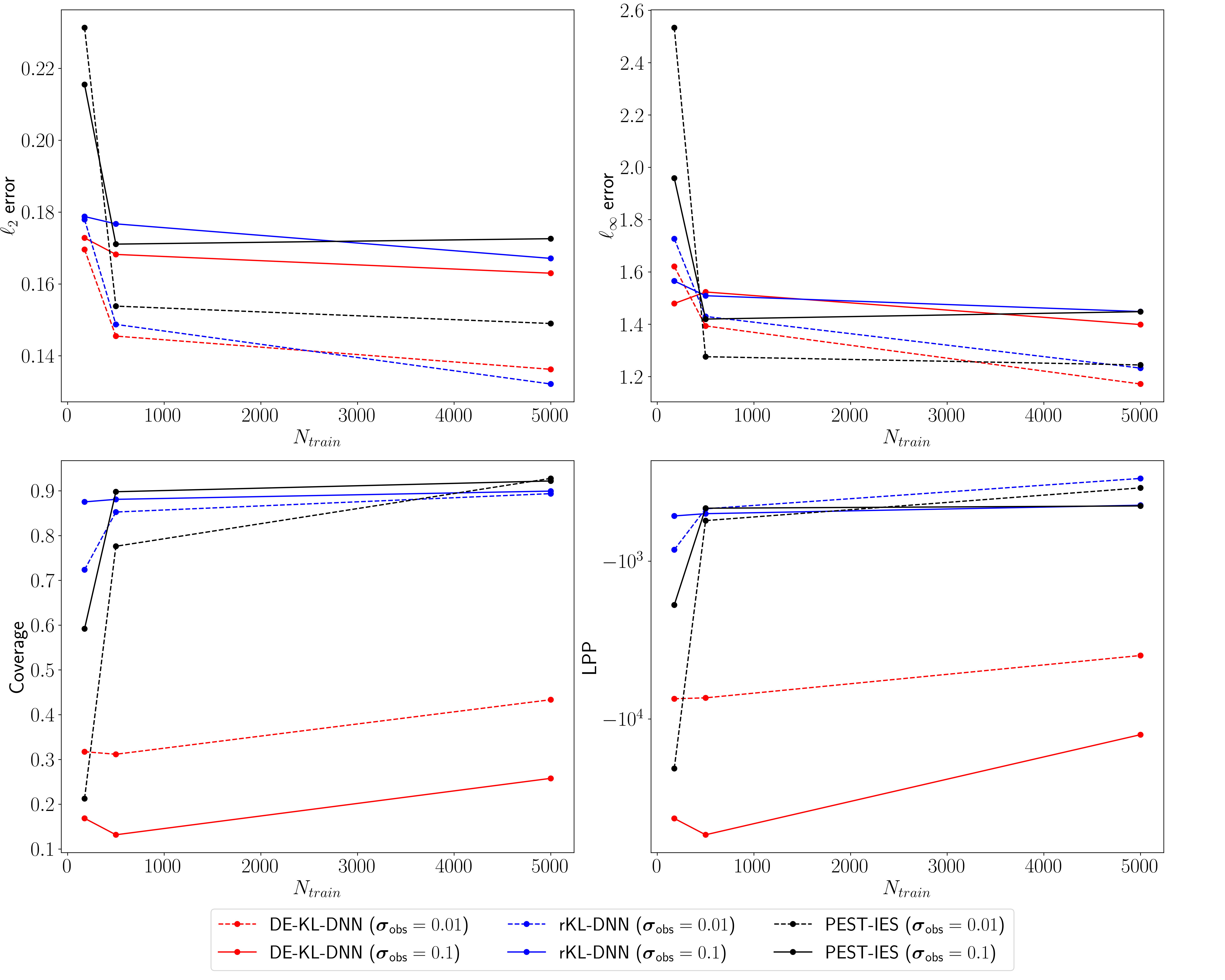}
    \caption{The $\ell_2$ and $\ell_\infty$ errors, coverages and LPPs of $y(\bm{x})$ estimated by DE KL-DNN (red lines), rKL-DNN (blue lines), IES (black lines) for $\sigma^2_{y^s} = 10^{-4}$  (dashed lines) and $10^{-2}$ (solid lines).}
    \label{fig:2d_est_error}
\end{figure}

\begin{table}[H]
    \centering
    \caption{The accuracy of $y$ estimates obtained with the DE-KL-DNN, rI-KL-DNN, and IES methods. The $\ell_2$ errors, LPPs, and coverages corresponding to $\sigma^2_{y^s}=10^{-4}$ and $10^{-2}$.  $N_{\text{train}}=5000$. }  
\begin{tabular}{l|lllll}
\hline$\sigma_{y^s}$ & Method & $\ell_2$ error & LPP & Coverage \\
\hline \multirow{3}{*}{$0.01$} & DE-KL-DNN & $0.136$ & $-3960.8$ & $306$ ($43.3 \%$)\\
& rI-KL-DNN & $0.132$ & $-299.3$ & $631$ ($89.4 \%$)\\
& PEST-IES & $0.149$ & $-343.5$ & $655$ ($92.8 \%$)\\
\hline \multirow{3}{*}{$0.1$} & DE-KL-DNN & $0.163$ & $-12569.2$ & $182$ ($25.8 \%$)\\
& rI-KL-DNN & $0.167$ & $-442.4$ & $635$  ($89.9 \%$)\\
& PEST-IES & $0.173$ & $-446.9$ & $651$ ($92.2 \%$)\\
\hline
\end{tabular}
\label{tab:estimated_y}
\end{table}

Figures \ref{fig:2d_diffu_coverage_0_1} and \ref{fig:2d_diffu_coverage_0_01} show
the means and standard deviations of the estimated $y(\bm{x})$ and the coverages (maps showing where $y_\text{ref}(\bm{x})$  is inside the predicted confidence intervals) obtained with the rI-KL-DNN, DE-KL-DNN, and IES methods for $\sigma^2_{y^s} = 10^{-2}$ and $10^{-4}$, respectively. The values of $\ell_2$ errors in the $y$ estimate given by the posterior mean, LPPs, and the percentages of coverage are given in Table \ref{tab:estimated_y}. We find that the three methods give similar mean estimates of $y(\bm{x})$ for both $\sigma^2_{y^s}$ values. The rI-KL-DNN and IES methods predict the posterior $y$ variances of the same order of magnitude, while the DE-KL-DNN method predicts an order of magnitude smaller variances. The coverages and LPPs are also similar in the rI-KL-DNN and IES methods but are significantly lower in the DE-KL-DNN method. As expected, the $y$ estimates obtained with less noisy measurements are more accurate (have smaller $\ell_2$ errors) and more descriptive (have larger LPP) than those obtained with noisier measurements. For $\sigma_{y^s}^2=0.01$, the DE method has the smallest $\ell_2$ errors and IES has the largest error. For $\sigma_{y^s}^2=10^{-4}$, the error is smallest in rI-KL-DNN and largest in IES. rI-KL-DNN has the largest LPP and DE-KL-DNN has the lowest LPP for both values of $\sigma_{y^s}^2$.

Next, we study how the quality of the $y$ estimates in the three methods depends on $N_\text{train}$. Figure \ref{fig:2d_est_error} shows the $\ell_2$ and $\ell_\infty$ errors, coverages, and LPPs as functions of $N_\text{train}$ for $\sigma_{y^s}^2 = 10^{-4}$ and $10^-2$. In general, the errors decrease and the coverages and LPPs increase with increasing $N_\text{train}$. The exception is IES where for the larger $\sigma_{y^s}^2$, the errors slightly increase with  $N_\text{train}$.  For all $N_\text{train}$ values, the $\ell_2$ errors in the DE-KL-DNN and rI-KL-DNN methods are similar and smaller than in the IES method. The $\ell_\infty$ errors are similar in the three methods for $N_\text{train}\ge 500$, but for smaller $N_\text{train}$, the DE-KL-DNN and rI-KL-DNN $\ell_\infty$ errors are 50\% smaller than in the IES method.  Overall, rI-KL-DNN and DE-KL-DNN provide the highest and lowest LPPs, respectively.  
The IES method exhibits the highest coverage, while the DE-KL-DNN method has the lowest coverage.
In all methods, the errors are smaller and the coverages and LPPs are larger for smaller $\sigma_{y^s}^2$.  
Based on these results, we conclude that the rI-KL-DNN method yields the most informative posterior distributions, whereas the DE-KL-DNN method produces the least informative posterior distributions.
Among the three methods, the performance of the IES method is most negatively affected by the small training (ensemble) size.

\subsection{Uncertainty in the predicted and forecasted $u$ fields}

In this section, we study the ability of the $y$ fields estimated with different methods to predict $u_\text{ref}$.
 We also study the ability of the estimated $y$ fields to forecast $u$ under arbitrary conditions. 
 Specifically, we forecast $u$ for a scenario where the pumping rates are increased by 50\% and the recharge rate is reduced by 40\% compared to those used for computing $u_\text{ref}$.
 Here, we compute $u$ using MODFLOW.  

Figure \ref{fig:2d_est_error} shows the $\ell_2$ and $\ell_\infty$ errors, coverages, and LPPs in the predicted $u_\text{ref}$ field as functions of $N_\text{train}$. Table \ref{tab:2d_est_error} lists the values of these variables for $N_{train}=5000$.
The errors are of the same order of magnitude in the three methods.
The accuracy of the rI-KL-DNN-based predictions increases with increasing $N_\text{train}$. This trend is reversed for the DE-KL-DNN method. The IES-based estimate is less sensitive to $N_\text{train}$.  
However, for  $N_\text{train} =100$, rI-KL-DNN has an order-of-magnitude larger LPP and 2-3 times larger coverage than the IES method. LPPs and coverages are similar in rI-KL-DNN and IES for $N_\text{train} \ge 500$.   
\begin{figure}[H]
    \centering
    \includegraphics[width=0.90\textwidth]{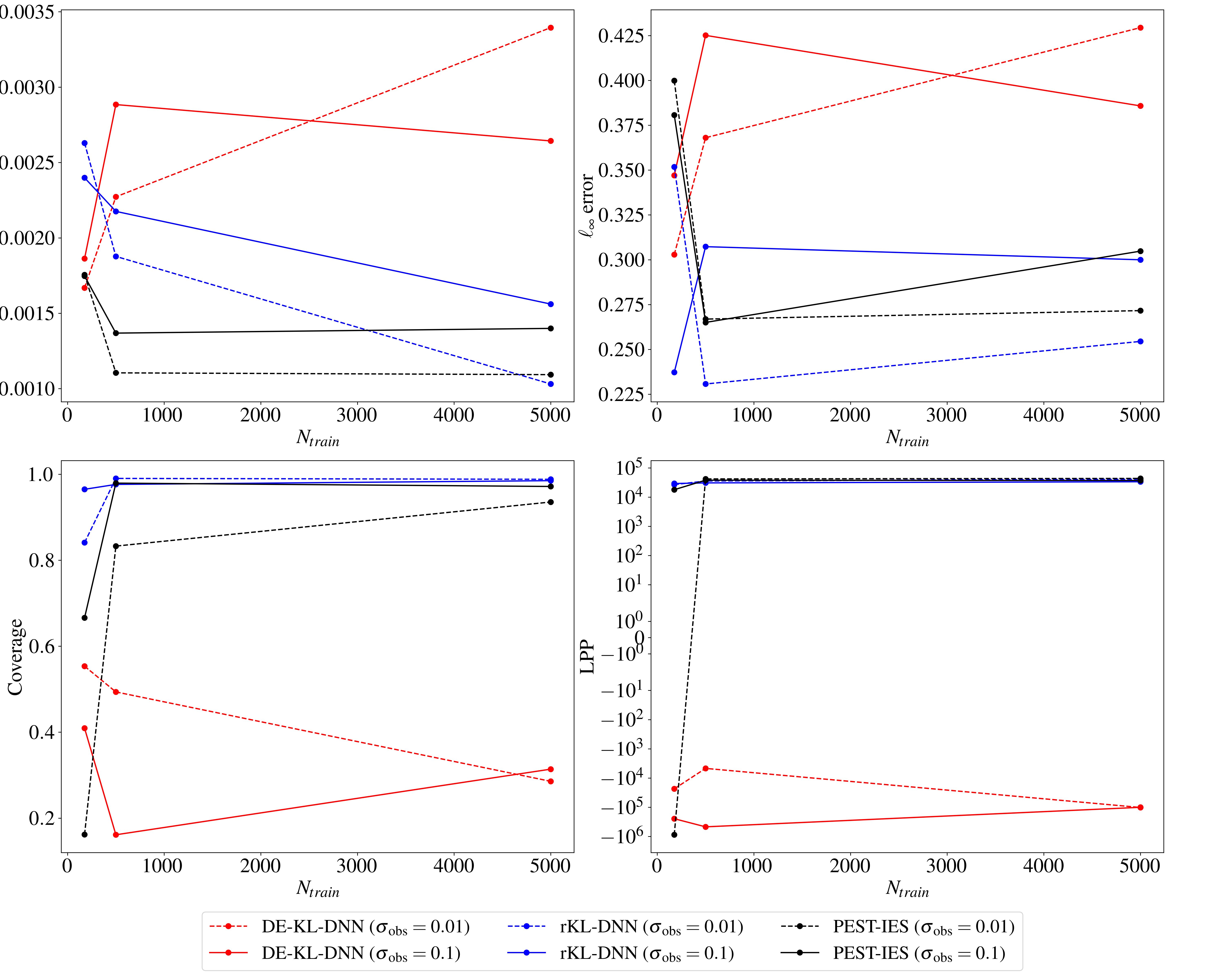}
    \caption{The accuracy of the estimated $y(\bm{x})$ in predicting $u_\text{ref}(\bm{x},t)$ as a function of $N_\text{train}$.
    The  $\ell_2$ and $\ell_\infty$ errors, coverage, and LPPs. The  $y(\bm{x})$  fields are estimated using noisy $u$ measurements with $\sigma^2_{u^s} = 10^{-4}$ and $10^{-2}$. }
    \label{fig:2d_est_error}
\end{figure}

\begin{table}[H]
    \centering
    \caption{
    The $\ell_2$ and $\ell_2$ errors, LPPs, and coverages for the estimated $u_\text{ref}$.  The $y(\bm{x})$ fields are estimated with the DE-KL-DNN, rI-KL-DNN, and IES methods using $N_{\text{train}}=5000$ and the noisy measurements of $u$ with  $\sigma^2_{u^s} = 10^{-4}$ and $10^{-2}$.}
\begin{tabular}{l|lllll}
\hline$\bm{\sigma}_{y^s}$ & Method & $\ell_2$ error & $\ell_\infty$ error & LPP & Coverage \\
\hline \multirow{3}{*}{$0.01$} & DE-KL-DNN & $3.39\times 10^{-3}$ & $0.429$ & $-100616$ &$28.6 \%$\\
& rI-KL-DNN & $1.03\times 10^{-3}$ & $0.254$ & $41414$ &  $98.8\%$\\
& PEST-IES & $1.09\times 10^{-3}$ & $0.272$ & $43729$ & $93.5 \%$\\
\hline \multirow{3}{*}{$0.1$} & DE-KL-DNN & $2.64\times 10^{-3}$ & $0.386$ & $-100616$ & $31.4 \%$\\
& rI-KL-DNN & $1.56\times 10^{-3}$ & $0.300$ & $ 33599$ & $98.5 \%$\\
& PEST-IES & $1.40\times 10^{-3}$ & $0.305$ & $37115$ & $97.2 \%$\\

\hline
\end{tabular}
\label{tab:2d_est_error}
\end{table}

\begin{figure}[H]
    \centering
    \includegraphics[width=0.90\textwidth]{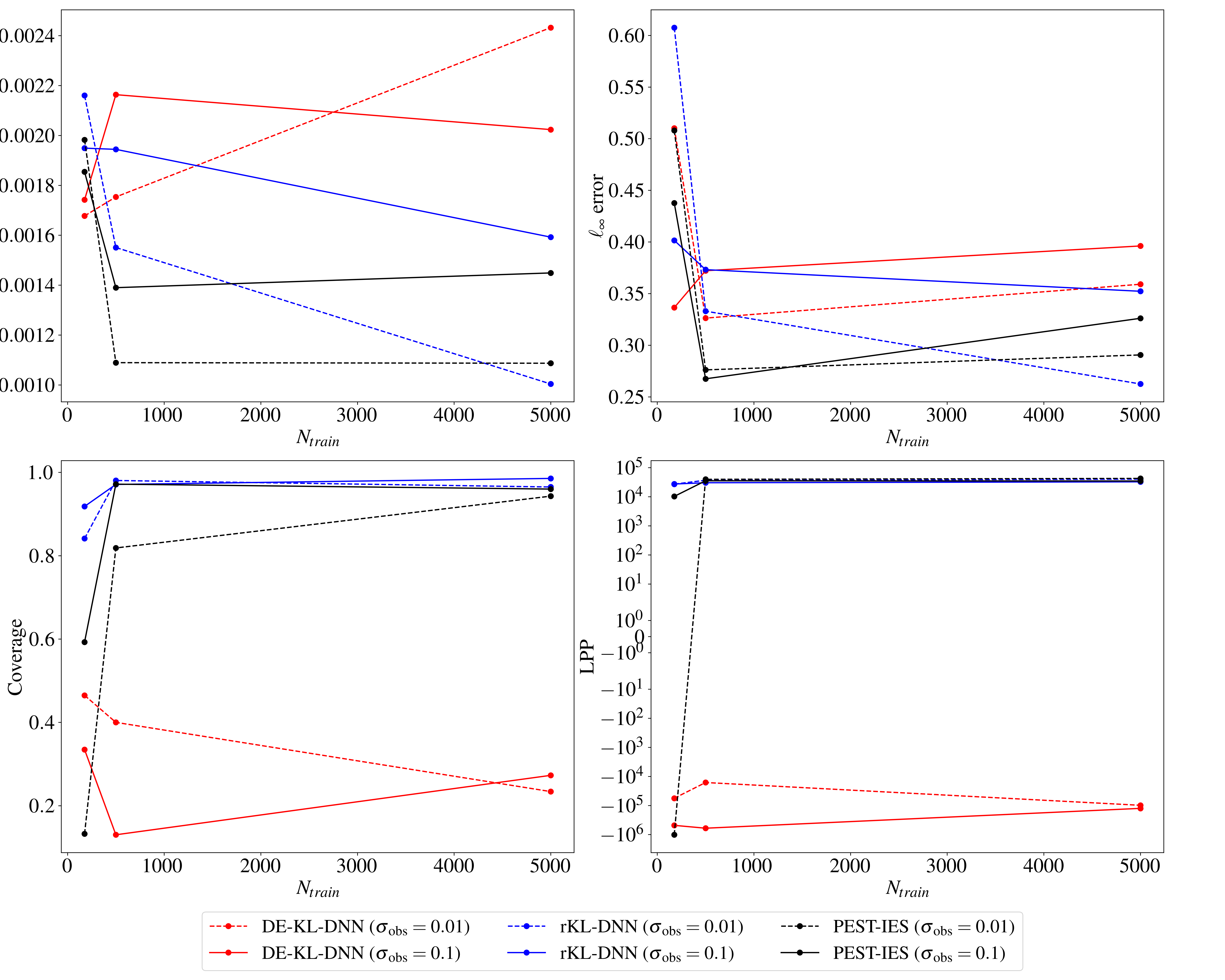}
    \caption{Forecast of $u(\bm{x},t)$ with the estimated $y(\bm{x})$. The $\ell_2$ and $\ell_\infty$ errors, coverages, and LPPs as functions of $N_\text{train}$ in the forecasted $u(\bm{x},t)$. 
    The $y(\bm{x})$ fields are estimated with the DE-KL-DNN, rI-KL-DNN, and IES methods using various $N_{\text{train}}$ values and the noisy measurements of $u$ with  $\sigma^2_{u^s} = 10^{-4}$ and $10^{-2}$.
}
    \label{fig:2d_forc_error}
\end{figure}

Figure \ref{fig:2d_forc_error} and Table \ref{tab:2d_forc_error} show the accuracy of the $u$ forecast obtained with $y$ estimated from the three methods for different $N_{train}$ and $\sigma^2_{y^s}$. The three methods have similar performance for forecasting and prediction. For $N_{train}=100$, the rI-KL-DNN-based forecast is more informative with larger LPPs and coverages than those in IES. The DE-KL-DNN method produces overall lower coverages and LPPs than the other two methods, except for $N_{train}=100$ when the DE LPPs and coverages are similar to those in IES. These results lead to the following conclusions: rI-KL-DNN produces a more informative $u$ prediction and forecast than the IES method for small training datasets and similar prediction and forecast for larger datasets. Both methods outperform DE-KL-DNN for UQ in predicting and forecasting $u$.

\begin{table}[H]
    \centering
    \caption{
        The $u(\bm{x},t)$ forecast using the $y(\bm{x})$ fields are estimated with the DE-KL-DNN, rI-KL-DNN, and IES methods  
    The $\ell_2$ and $\ell_2$ errors, LPPs, and coverages for forecasted $u$. $N_{\text{train}}=5000$ and the noisy measurements of $u$ with  $\sigma^2_{u^s} = 10^{-4}$ and $10^{-2}$.
    }
\begin{tabular}{l|lllll}
\hline$\bm{\sigma}_{y^s}$ & Method & $\ell_2$ error & $\ell_\infty$ error & LPP & Coverage \\
\hline \multirow{3}{*}{$0.01$} & DE-KL-DNN & $2.43\times 10^{-3}$ & $0.359$ & $-98450$ &$23.4 \%$\\
& rI-KL-DNN & $1.00\times 10^{-3}$ & $ 0.262$ & $40441$ &  $96.5 \%$\\
& PEST-IES & $1.09\times 10^{-3}$ & $0.291$ & $42736$ & $94.3 \%$\\
\hline \multirow{3}{*}{$0.1$} & DE-KL-DNN & $2.03\times 10^{-3}$ & $0.396$ & $-125506$ & $27.3 \%$\\
& rI-KL-DNN & $1.59\times 10^{-3}$ & $0.352$ & $32601$ & $98.5 \%$\\
& PEST-IES & $1.44\times 10^{-3}$ & $0.326$ & $34772$ & $96.0 \%$\\

\hline
\end{tabular}
\label{tab:2d_forc_error}
\end{table}

\section{Conclusions}\label{sec:conclusions}
We proposed the rI-KL-DNN method, an approximate likelihood-free method for quantifying total uncertainty in inverse PDE solutions by sampling the posterior distribution of the parameters. This method enables Bayesian data assimilation in high-dimensional problems using a surrogate model while accounting for uncertainty in the surrogate model and traditional sources of uncertainty due to the PDE model and measurement errors. The rI-KL-DNN method allows for non-Gaussian surrogate model errors. In this work, we used the KL-DNN reduced-order deep learning surrogate model, and the samples of the error distribution are obtained with the rKL-DNN algorithm. In this problem, the likelihood evaluation requires marginalization over the surrogate model parameters, which is computationally unfeasible. Therefore, the sampling methods requiring likelihood evaluations, such as MCMC,  cannot be directly used for this problem.  The proposed method is compared with the DE and IES methods for the inverse non-linear PDE problem describing groundwater flow in a synthetic unconfined aquifer. 

We found that the rI-KL-DNN method produces more informative posterior distributions of the parameters and states (larger LPPs and coverage) than the DE-KL-DNN method, which only accounts for uncertainty due to DNN training in the surrogate model.  The comparison with the IES method revealed that the rI-KL-DNN yields more informative parameter and state distributions for small training datasets (ensemble sizes in IES) and similar predictions for large training datasets. Our results show that despite inherent uncertainty, surrogate models can be used for parameter and state estimation as an alternative to the inverse methods relying on (more accurate) numerical PDE solvers.

\section{Acknowledgements}
The authors are thankful to James L. McCreight and Joseph D. Hughes for their help with generating training datasets and performing MF6 and PEST++ simulations  
This research was partially supported by the U.S. Geological Survey under Grant No. G22AP00361, the DOE project ``Science-Informed Machine Learning to Accelerate Real-time (SMART) Decisions in Subsurface Applications Phase 2 – Development and Field Validation,''  the U.S. Department of Energy (DOE) Advanced Scientific Computing program, and the United States National Science Foundation. Pacific Northwest National Laboratory is operated by Battelle for the DOE under Contract DE-AC05-76RL01830. 

The views and conclusions contained in this work are those of the authors and should not be interpreted as representing the opinions or policies of the U.S. Geological Survey. Mention of trade names or commercial products does not constitute their endorsement by the U.S. Geological Survey.

\bibliographystyle{elsarticle-num}
\bibliography{sample}

\end{document}